\documentclass[10pt,twocolumn,letterpaper]{article}
\pdfoutput=1

\usepackage[utf8]{inputenc}
\usepackage{iccv}
\usepackage{times}
\usepackage{epsfig}
\usepackage{graphicx,subfigure}
\usepackage{amsmath}
\usepackage{amssymb}
\usepackage{makecell}
\usepackage{arydshln}
\usepackage{gensymb}
\usepackage{pdfpages}
\usepackage{booktabs}
\usepackage{stmaryrd}

\newcommand{\ra}[1]{\renewcommand{\arraystretch}{#1}}

\usepackage{multirow}
\usepackage{array}
\newcolumntype{?}{!{\vrule width 1pt}}
\usepackage[square,numbers,sort]{natbib}

\usepackage{xcolor}

\newcommand{\commented}[1]{}

\newcommand{\vincentrmk}[1]{}

\newcommand{\hugormk}[1]{}

\usepackage[pagebackref=true,breaklinks=true,urlcolor=blue,letterpaper=true,colorlinks,bookmarks=false]{hyperref}

\iccvfinalcopy 

\def\iccvPaperID{1013} 

\ificcvfinal\fi
\begin{document}

\title{Improving Nighttime Retrieval-Based Localization}
\author{Hugo Germain\footnotemark[1]~~\textsuperscript{1} \hspace{1em}
Guillaume Bourmaud\footnotemark[1]~~\textsuperscript{2} \hspace{1em}
Vincent Lepetit\footnotemark[1]~~\textsuperscript{1} \\ \textsuperscript{1}Laboratoire Bordelais de Recherche en Informatique, Université de Bordeaux, France\\ \textsuperscript{2}Laboratoire IMS, Université de Bordeaux, France\\}
\maketitle
\setcounter{footnote}{1}\footnotetext{E-mail: {\tt\small \{firstname.lastname\}@u-bordeaux.fr}}


\begin{abstract}

Outdoor visual localization is a crucial component to many computer vision systems.  We propose an approach to localization from images that is designed to explicitly handle the strong variations in appearance happening between daytime and nighttime.  As revealed by recent long-term localization benchmarks, both traditional feature-based and retrieval-based approaches still struggle to handle such changes.  Our novel localization method combines a state-of-the-art image retrieval architecture with condition-specific sub-networks allowing the computation of global image descriptors that are explicitly dependent of the capturing conditions.  We show that our approach improves localization by a factor of almost 300\% compared to the popular VLAD-based methods on nighttime localization.

 \end{abstract}

\section{Introduction}

Outdoor environments are prone to large changes in visual appearance, both throughout the time of day and across longer periods of time. Such variations strongly impact the performance of image-based algorithms, for which visual localization is no exception. The goal of visual localization is to predict the 6 DoF camera pose for a visual query  with respect to a reference frame. It is a crucial component of many robotic systems, including popular research areas like autonomous navigation~\cite{McManus2014ShadyDR} and augmented reality~\cite{Middelberg2014Scalable6L}.

\begin{figure}[t]
\begin{center}
\includegraphics[width=0.95\linewidth]{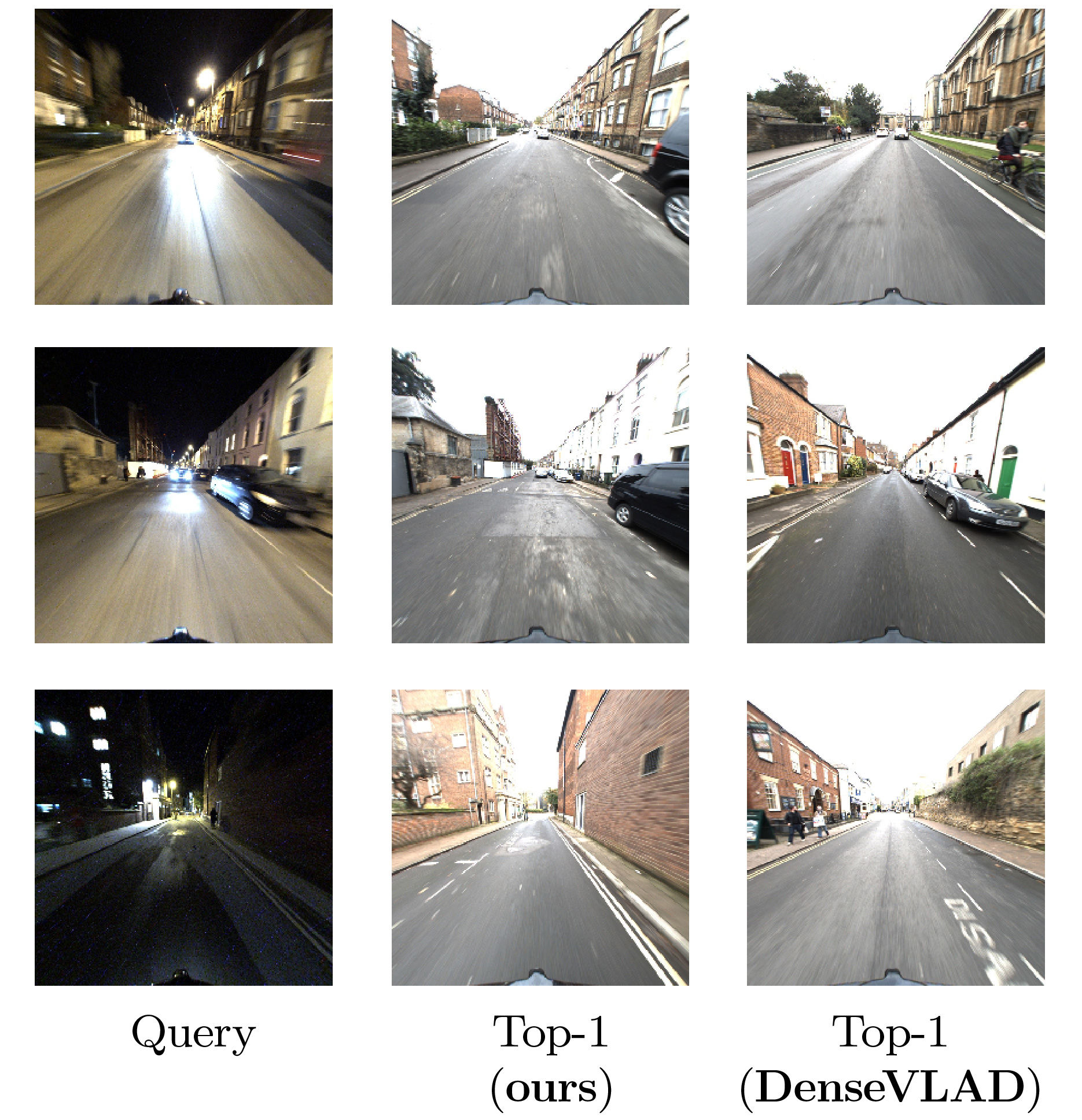}
\end{center}
\vspace{0.25cm}
 \caption{Overview of our approach: Given a pose-annotated image
   database captured in daytime only, our goal is to accurately
   relocalize query images captured in nighttime
   conditions 
    within this database. We introduce a simple and efficient architecture which explicitly considers the capturing conditions of the query images to compute their descriptors, in order to localize them by retrieval. Our method (middle column) increases localization results at nighttime compared to traditionally used VLAD-based methods~\cite{Torii2015247PR, Arandjelovix01072016NetVLADCA} (right) by up to 42\% on night images from RobotCar Seasons~\cite{sattler:hal-01859660}.}\label{fig:intro}
\label{fig:long}
\label{fig:onecol}
\end{figure}

Visual  localization  methods  usually  fall  into  either  of  two  categories:
\textit{Feature-based}  methods that  leverage  a  3D model  of  the scene,  and
\textit{retrieval-based} methods  that infer the  camera pose by  retrieving the
nearest image(s)  in a  geo-tagged database~\cite{Arandjelovic2014DisLocationSD,
  Chen2011CityscaleLI,  Torii2015247PR,  Arandjelovix01072016NetVLADCA}.   While
inducing approximation  in pose estimation, retrieval-based  localization offers
two key advantages over feature-based localization: First, it is easily scalable
to large environments. Only the compact  image descriptors, which can further be
compressed  with quantization  techniques~\cite{Jgou2011ProductQF},  need to  be
stored while  feature-based methods need  to build and store  a 3D model  of the
scene.   Second, thanks  to  the image  descriptors robustness,  retrieval-based
localization   currently  outperforms   traditional  feature-based   methods  in
difficult  capturing conditions (such  as  over-exposed sunny
images)    at   a   coarse   level   of  precision,   as   shown   by   recent
benchmarks~\cite{sattler:hal-01859660}.    These   two  advantages   make
retrieval-based approach a suitable candidate for long-term visual localization.

In this paper, we consider the case where the images of the geo-tagged
database were captured under daytime conditions while visual queries were taken at night. This is a scenario very important in practice: For example,
the database images could be all captured during the same day, but
should still be used to localize new images captured at night.

Despite the recent progress in learning-based methods for localization,
the overall  performance of  current retrieval-based localization approaches~\cite{Torii2015247PR, Arandjelovix01072016NetVLADCA} yields unsatisfactory results, and nighttime visual localization is still far
  from being solved,  as Figure~\ref{fig:intro} shows.  In this  paper, we argue
  that this limitation in performances is due to the fact that:
\begin{itemize}
\item Modern deep retrieval architectures have  not been employed for long-term visual
localization yet. Recent architectures~\cite{Gordo2017EndtoEndLO, Radenovic2018RevisitingOA, Radenovic2018FinetuningCI} have shown to outperform VLAD-based representations~\cite{Torii2015247PR, Arandjelovix01072016NetVLADCA,
DBLP:journals/corr/abs-1809-09767} for the task of image retrieval, and could therefore be employed for retrieval-based localization, especially for the difficult nighttime scenario.
\item The lack of training data at nighttime makes it very difficult to efficiently leverage
deep architectures without over-fitting. As our experiments will show, adapting
  the way the image descriptor is computed with respect to the capturing condition
  can result in significant performance improvement.
\end{itemize}


Thus, we make the following contributions:
\begin{itemize}
\item We experimentally demonstrate that using a state-of-the-art image retrieval pooling layer~\cite{Radenovic2018FinetuningCI} with a siamese architecture allows to outperform both feature-based and retrieval-based methods by a  large  margin  in  the  case  of  coarse  nighttime  localization, where it is difficult to detect and match repeatable feature points.

\item To further improve the localization results under nighttime conditions, we propose to make the computation of the image descriptors explicitly dependent of the capturing conditions of the input image by introducing condition-specific sub-networks. These sub-networks allow us to adapt
to the variation of landmark appearances due to the conditions.

\end{itemize}

The paper is structured as follows: Section~\ref{relatedwork} reviews related work. Section~\ref{method} introduces our novel localization pipeline. Section~\ref{experiments} describes our experimental setup to thoroughly evaluate our approach in the context of long-term localization. Finally, Section~\ref{results} shows our localization results.

\section{Related Work}\label{relatedwork}

Visual localization methods usually fall into either of two categories: those leveraging a 3D model of the scene~(\textit{feature-based}), and those inferring the camera pose by retrieving the nearest image(s) in an annotated corpus~(\textit{retrieval-based}). We briefly review both approaches below in the context of long-term localization.

\begin{figure*}
\begin{center}
\includegraphics[width=0.95\linewidth]{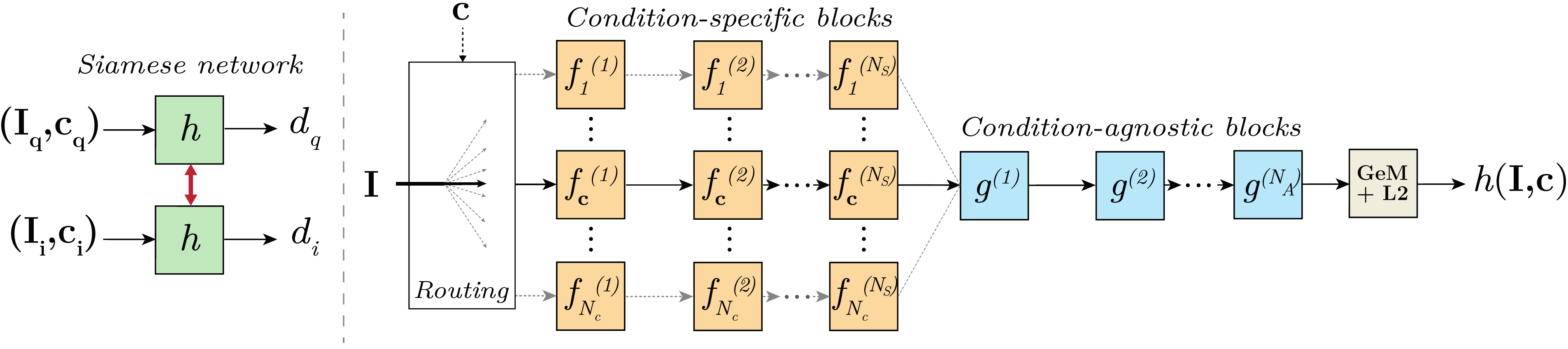}
\end{center}
\vspace{0.15cm}
\caption{Detailed view of our siamese architecture. At a high level, our network is similar to a standard image-retrieval network, followed by a Generalized-Mean-Pooling~(GeM)~\cite{Radenovic2018FinetuningCI} activation and L2-normalization (see left). To exploit the image capturing condition, our network consists of $N_S$ condition-specific blocks $(f_c^{(1)}, ..., f_c^{(N_S)})$ for every condition, followed by $N_A$ condition-agnostic blocks $(g^{(1)}, ..., g^{(N_A)})$ (see right). When computing global image descriptors, inputs are routed to their appropriate condition-expert backbone. Note that the computational cost at inference time is independent of $N_c$, and the only additional cost is in terms of memory.}
\label{fig:short}\label{fig:detailed}
\end{figure*}

\subsection{Feature-based localization}
Traditional visual localization methods \cite{Li2012WorldwidePE,Li:2010:LRU:1888028.1888088,Liu2017EfficientG2,Sattler2015HyperpointsAF,sattler:hal-01513083,svarm_csl} directly regress the full 6~DoF camera pose of query images, with respect to a known reference. A typical pipeline of such feature-based methods includes acquiring a point-cloud model of the scene through \textit{SfM}, and computing local feature descriptors like SIFT~\cite{Lowe:2004:DIF:993451.996342} or LIFT~\cite{Yi2016LIFTLI}. These descriptors are in turn used to establish 2D-to-3D correspondences with every point of the model, and the query's camera pose can be directly inferred from those matches, using RANSAC~\cite{Fischler1981RandomSC, Sattler2014OnSF} combined with an  Perspective-n-Point~(P\textit{n}P) solver~\cite{Bujnak2010NewES, Haralick1994ReviewAA, Kukelova2013RealTimeST}.

Feature-based methods achieve very competitive results in consistent daytime conditions~\cite{Sattler2017EfficientE, svarm_csl, sattler:hal-01859660, sattler:hal-01513083, Walch2017ImageBasedLU}. However, the pose estimation accuracy for such methods is heavily reliant on the local 2D-3D correspondences. Research in feature-based approaches mostly focuses on improving descriptor matching efficiency~\cite{Choudhary:2012:VPS:2403138.2403150,Li:2010:LRU:1888028.1888088,Sattler2017EfficientE, Lim2012RealtimeI6, Larsson2016OutlierRF, Lynen2015GetOO}, speed~\cite{Heisterklaus2014ImagebasedPE,Donoser2014DiscriminativeFM} and robustness~\cite{Svrm2014AccurateLA,Sattler2016LargeScaleLR, Li2012WorldwidePE, Sattler2015HyperpointsAF, svarm_csl, Zeisl2015CameraPV}. Despite these efforts, a significant number of erroneous matches appear under strong appearance variations. As a result, dramatic drops in localization accuracy are observed in harder conditions such as nighttime~\cite{sattler:hal-01859660}.

One workaround is to avoid using explicit feature matching altogether, and train a CNN to regress dense scene coordinates~\cite{Brachmann2017LearningLI,Bui2018SceneCA,Kendall2017GeometricLF}. A confidence-ranking scheme such as DSAC~\cite{Brachmann2017DSACD} is then used to discard erroneous pose candidates sampled from the obtained depth map. However, such learning-based variants are hard to initialize~\cite{sattler:hal-01859660, Schnberger2017SemanticVL} and struggle with large environments~\cite{sattler:hal-01859660}. Other direct learning-based variants include end-to-end pose regression from images~\cite{Kendall2015PoseNetAC, Walch2017ImageBasedLU}, although these approaches are overall less accurate~\cite{sattler:hal-01859660, Brachmann2017LearningLI, Balntas_2018_ECCV}.

Another way of making feature-based methods robust to condition perturbations is to use semantic reasoning. Semantic information can indeed be exploited to enhance either the feature matching stage~\cite{Arandjelovic2014VisualVW,Kobyshev2014MatchingFC,pub.1046137732, Schnberger2017SemanticVL} or the pose estimation stage~\cite{Toft_2018_ECCV}. While being accurate at small scale, feature-based methods bottleneck is scalability. Both the construction of precise 3D models~(and their maintenance), and local feature-matching is challenging and expensive in large-scale  scenarios~\cite{sattler:hal-01513083}.

\subsection{Retrieval-based localization}
Retrieval-based or image-based localization methods trade-off accuracy for scalability, by modeling the scene as an image database, and visual localization as an image retrieval problem. Visual queries are matched with a pose-annotated images, using compact image-level representations. The query's camera pose can then be simply inferred from the top-ranked images~\cite{Chen2011CityscaleLI,Zamir2010AccurateIL,Zhang2006ImageBL,sattler:hal-01513083}, and the need for ground-truth 3D geometry is alleviated.

Robust global descriptors can be obtained by aggregation of local features in the image. VLAD~\cite{Arandjelovic2013AllAV} is a popular descriptor, computed by summing and concatenating many descriptors for affine-invariant regions. DenseVLAD~\cite{Torii2015247PR} modified the VLAD architecture by densely sampling RootSIFT~\cite{Arandjelovic2012ThreeTE} descriptors in the image. Recent learning-based variants cast the task of image retrieval as a metric learning problem. NetVLAD~\cite{Arandjelovix01072016NetVLADCA} defines a differentiable VLAD layer as the final activation of a siamese network. While never explicitly applied to visual localization, activations such as max-pooling~(MAC)~\cite{Tolias2015ParticularOR,Razavian2014VisualIR}, sum-pooling~(SPoC)~\cite{Babenko2015AggregatingLD}, weighted sum-pooling~(CroW)~\cite{Kalantidis2016CrossdimensionalWF} or regional max-pooling~(R-MAC)~\cite{Gordo2017EndtoEndLO} coupled with siamese or triplet architectures, have shown to outperform VLAD-based methods in image-retrieval tasks~\cite{Radenovic2018RevisitingOA}. The Generalized-Mean-Pooling~(GeM)~\cite{Radenovic2018FinetuningCI} layer proposes a hybrid activation, which combines the benefits of average and max-pooling, and achieves state-the-art results in outdoor visual search~\cite{Radenovic2018RevisitingOA}.

One bottleneck in retrieval-based localization is the spatially-sparse image sampling of the database. Three schemes can be implemented to compensate for the induced pose approximation: i) View synthesis~\cite{Torii2015247PR} artificially generates intermediate samples, ii) relative pose regression~\cite{Taira2018InLocIV, Balntas_2018_ECCV} acts as a separate refinement step, iii) multi-image methods~\cite{Zamir2010AccurateIL, Zhang2006ImageBL, Balntas_2018_ECCV} combine the top ranked images to improve pose accuracy.

Compared to 3D feature-based methods, retrieval-based methods offer two key advantages. First, the extension of such methods to city-scale scenarios is trivial~\cite{sattler:hal-01513083}. Besides, in a very large database, unsupervised descriptor compression like PCA~\cite{Jgou2012NegativeEA} or Product Quantization~(PQ)~\cite{Jgou2011ProductQF} enables efficient approximate nearest-neighbour search with little loss in performance~\cite{Gordo2017EndtoEndLO}. Secondly, evaluations in long-term~\cite{sattler:hal-01859660,Toft_2018_ECCV, Arandjelovix01072016NetVLADCA, Torii2015247PR} visual localization reveal that retrieval-based methods outperform other feature-based methods like Active Search~\cite{Sattler2017EfficientE} at a coarse precision level. Image pre-processing~\cite{McManus2014ShadyDR} or translation~\cite{DBLP:journals/corr/abs-1809-09767} can be employed to bring all images to a visually similar and condition-invariant representation. However, image translation methods are complex to train, require a full retraining for every condition as well as a large amount of images. They also add to the overall computational cost of retrieval methods. Our method outperforms both retrieval and feature-based techniques in the case of coarse nighttime localization without the need for such pre-processing.


\section{Method}\label{method}

In this section, we give an overview of our pipeline.

\subsection{Problem statement}
As discussed in the introduction, we would like to train a method to compute a descriptor for a given image in a way that depends on the capturing conditions with the objective of improving nighttime localization results. We exploit a training set made of images annotated with the 3D pose for the camera, and the capturing conditions:
\begin{equation}
    \mathcal{D} = \{(I_i, c_i, M_i) \forall i  \in (1, …, N) \} \> ,
\end{equation}
where $c_i \in \mathcal{C}$ is the capturing condition for $I_i$, and $M_i$ is its camera pose.
We assume a set $\mathcal{C}$ of $N_c$ finite and discrete capturing conditions, \ie  $\mathcal{C}=\{c_1,...,c_{N_c}\}$. In practice, we use the following set of conditions: Rain, dawn, snow, dusk, sun, overcast in summer and winter, as well as night and rain at nighttime (\textit{night-rain}). Even if we are mainly interested in improving nighttime localization, we consider all available capturing conditions in order to prevent over-fitting.
For supervised training purposes we define positive and negative labels $l(I_i,I_j)\in\{0,1\}$, depending if the camera poses for $I_i$ and $I_j$ present visual overlap or not. Details regarding those steps are provided in Section~\ref{mining}.

\subsection{Architecture}
Many previous methods leverage a siamese or triplet network~\cite{Tolias2015ParticularOR,Razavian2014VisualIR, Arandjelovix01072016NetVLADCA, Kalantidis2016CrossdimensionalWF, Gordo2017EndtoEndLO, Radenovic2018FinetuningCI} to perform image retrieval. Like the recent
\cite{Radenovic2018FinetuningCI}, we opt for a siamese architecture, mostly for its simplicity. The architecture of our network is presented in Figure~\ref{fig:detailed}.

\noindent\textbf{Learning condition-based representations}
As shown in Figure~\ref{fig:detailed}, a key difference with previous approaches is that our architecture explicitly introduces the capturing condition: When dealing with multiple conditions, traditional methods define a single network to learn a joint representation across all capturing condition. While simple and compact, we argue that this approach struggles to handle strong appearance variations when given little training images per condition. Instead, we compute our image descriptor $h(I, c)$ for a given image $I$ and its condition $c$ with a network made of three components:
\begin{equation}
    h(I, c;\Theta_1,\ldots\Theta_{N_c},\Phi) = G\Big(g\Big(\prod_{i=1}^{N_c}{f_i(I;\Theta_i)^{\delta(c-i)}};\Phi\Big)\Big)\>.
    \label{eq:h}
\end{equation}
where $\delta(\cdot)$ is the Dirac function.
We have therefore $N_c$ condition-specific sub-networks $(f_1,f_2,..,f_{N_c})$ of parameters $(\Theta_1,\Theta_2,..,\Theta_{N_c})$. \commented{Each network $f_i$ is condition-specific and takes an image as input.} Let us emphasize that during the forward pass, only the sub-network $f_c$ corresponding to the condition $c$ of the input image $I$ is executed, \ie in eq.(\ref{eq:h}) the Dirac function acts as a router. A second condition-agnostic sub-network $g$ of parameters $\Phi$ takes the output of the network $f_c$ as input. Function $G$ denotes Generalized-Mean Pooling~(GeM)~\cite{Radenovic2018FinetuningCI} followed by L2 normalization. It is applied to the output of $g$ to finally return  a descriptor. This Generalized-Mean Pooling is discussed below.

$G$ is parameter-free, and we train $g$ and the $f_c$'s jointly by minimizing a contrastive loss function as in \cite{Chopra2005LearningAS}:
\begin{equation}
\min_{\{\Theta_c\}_c,\Phi} \sum_{(I_i, c_i, I_j, c_j)} \mathcal{L}(I_i, c_i, I_j, c_j) \> ,
\end{equation}
where
\begin{equation}
\begin{array}{l}
    \text{$\mathcal{L}(I_i, I_j)$} = \\
    \begin{cases}{}
        ||h(I_i, c_i)-h(I_j, c_j)||_2^2 \>\> &\text{ if } l(I_i, I_j)=1 \\
        \max\{0, m-||h(I_i,c_i)-h(I_j,c_j)||_2\}^2 \>\>&\text{ otherwise, }\\
    \end{cases}
\end{array}
\end{equation}
where $m>0$ is a margin that can be set arbitrarily.

As a result of this optimization, the $f_c$ networks become experts at mapping their specific conditions to a shared representation space, \emph{with no additional computational cost at inference time compared to a standard pipeline}.

\noindent\textbf{Generalized-Mean Pooling and Normalization.} Final pooling layers of siamese networks are a key factor to generating robust and discriminative global image descriptors. The purpose of such layers is to efficiently aggregate local features into a fixed-length vector. Among existing techniques, the Generalized-Mean Pooling~(GeM)~\cite{Radenovic2018FinetuningCI} layer
generalizes average~\cite{Babenko2015AggregatingLD} and global maximum pooling~\cite{Razavian2014VisualIR, Tolias2015ParticularOR}, and
has led to state-of-the-art results in learning-based image retrieval. We thus use it for our function $G$ as the last component of $h$, together with a L2 normalization, instead of the more traditionally used VLAD-based representations~\cite{Arandjelovix01072016NetVLADCA, Torii2015247PR}.

More exactly, let $\mathcal{X}$ be a convolutional response of size $N \times M \times K$, and $\{\mathcal{X}_k, k \in (1,...,K)\}$ the set of $N \times M$ feature maps.
We take:
\begin{equation}
G(\mathcal{X}) = \frac{{\bf d}}{\|{\bf d}\|}, \text{ where }
\end{equation}
\begin{equation}
    {\bf d} = \begin{pmatrix} d_{1} \\ \vdots \\ d_{K} \end{pmatrix}
    \text{ with } d_k = \left(\frac{1}{|\mathcal{X}_k|}\sum_{x\in \mathcal{X}_k}{x^{p_k}}\right)^\frac{1}{p_k}
\end{equation}
$p_k$ acts as a factor between average ($p_k=1$) and max-pooling ($p_k\rightarrow\infty$).
The resulting global descriptors have proved to be highly discriminative and to deliver state-of-the-art results in challenging datasets~\cite{Radenovic2018RevisitingOA}. To the best of our knowledge, it is the first time that the GeM layer is applied in the context of visual localization.

\subsection{Localization by retrieval}\label{loc}
Finally, we perform localization by approximating the query pose with the one of the top-ranked image. We pre-compute normalized global descriptors for every image in the database using the aforementioned method, and find the nearest image by cosine similarity with the query's descriptor. To remove descriptor noise due to over-represented features, we learn and apply descriptor whitening discriminatively using our training data~\cite{Radenovic2018FinetuningCI}. More specifically, we subtract the mean GeM vector $\mu$ and rescale the descriptors using the covariance estimated from positive samples. Note that in the case of a very large database, the pose can be inferred efficiently using a combination of approximate nearest-neighbours and descriptor compression~\cite{Gordo2017EndtoEndLO, Jgou2012NegativeEA, Jgou2011ProductQF}.

\section{Experimental Setup}\label{experiments}
In this section, we conduct experiments to evaluate our approach in the context of nighttime visual localization.
\subsection{Dataset}
We train and evaluate our approach on the RobotCar-Seasons  dataset made by Sattler \etal~\cite{sattler:hal-01859660}, which is a revision of the Oxford RobotCar dataset~\cite{Maddern20171Y1}. The original Oxford RobotCar dataset consists of outdoor urban images captured from a car and over the course of a full year, across overlapping trajectories. It features both short-term (\eg day-to-night, weather) and long-term (\eg seasonal) environment evolutions, and therefore different conditions. Images were captured as triplets along three fixed viewpoints (left, rear, and right).

The RobotCar Seasons refines a subset of the original dataset by recomputing pose annotations more accurately (using LIDAR scans, INS, SfM and manual annotation), and by labelling images with discrete capturing conditions. The release of this new dataset came along with a new benchmark, aiming specifically at evaluating visual localization algorithms robustness to strong appearance changes.

More exactly, the RobotCar Seasons dataset is split in three separate groups. The first one, which we refer to as \textit{overcast-reference} and will play the role of the geo-tagged database at test-time, is made of 20,862 pose-annotated images captured in a single traversal during daytime. \textit{Overcast-reference} covers the whole scene and is provided with a corresponding 3D point cloud. The second one,  called \textit{mixed-conditions}, consists of 5,718 pose-annotated images equally distributed across 9 different capturing conditions, including 1,263 nighttime images.  Lastly, the \textit{query} group consists of 2,634 nighttime images, which overlap with the \textit{overcast-reference} traversal but not with \textit{mixed-conditions}. This makes our approach challenging because we cannot leverage cross-condition information for the reference traversal, and thus need to generalize appearance variations on spatially separate information. In our evaluation, we always make predictions using single images at a time and do not leverage associated triplets to infer the query camera pose.
Since we focus on nighttime localization, we assume the capturing condition to be known at test time. In the absence of ground-truth labeling, one could easily obtain such information using a simple image classifier.

\begin{figure}
\begin{center}
\includegraphics[width=1.0\columnwidth]{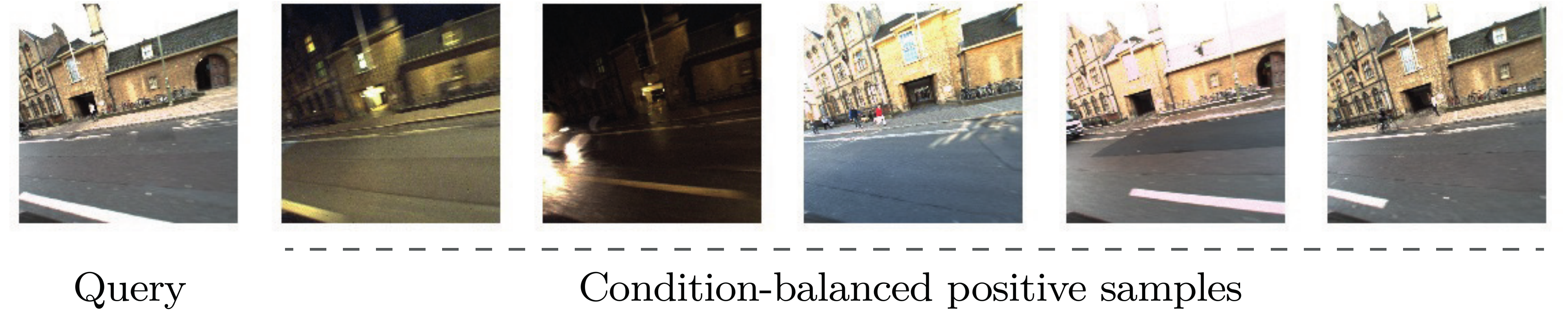}
\end{center}
\caption{Example query with its associated positive samples from \textit{mixed-conditions}, obtained through condition-balanced hard mining with relaxed inliers.}
\label{fig:short}\label{fig:sample}
\end{figure}

\begin{table}
\ra{1.3}
\begin{center}
\resizebox{\columnwidth}{!}{%
\begin{tabular}{@{}llcc@{}}
\toprule
$c_i$ & Conditions  & Number of images & Total per sub-branch \\
\midrule
$c_i=0$	& dawn & 690 & 690\\
\hline
$c_i=1$	& snow	& 717 & 717 \\
\hline
$c_i=2$	& sun	& 627 & 627  \\
\hline
\multirow{2}{*}{$c_i=3$} & \multicolumn{1}{l}{night} & \multicolumn{1}{c}{591} & \multirow{2}{*}{1,263}\\
                     & \multicolumn{1}{l}{night-rain} & \multicolumn{1}{c}{672} &\\
\hline
\multirow{3}{*}{$c_i=4$} & \multicolumn{1}{l}{overcast-summer} & \multicolumn{1}{c}{660} & \multirow{3}{*}{3,266}\\
                     & \multicolumn{1}{l}{overcast-winter} & \multicolumn{1}{c}{606} &\\
                     & \multicolumn{1}{l}{overcast-reference} & \multicolumn{1}{c}{20,862*} & \\
\hline
\multirow{2}{*}{$c_i=5$} & \multicolumn{1}{l}{dusk} & \multicolumn{1}{c}{561} & \multirow{2}{*}{1,155}\\
                     & \multicolumn{1}{l}{rain} & \multicolumn{1}{c}{594} &\\
\bottomrule
\end{tabular}%
}
\end{center}
\caption{Capturing condition binning of the training set. To avoid diluting images across too many sub-networks $f_c$, we merge some conditions based on visual similarity (*~Due to the relatively large number of images in \textit{overcast-reference}, 2,000 images from the 20,862 are resampled at every epoch).}\label{table:condition_binning}
\end{table}

\subsection{Training Set}\label{mining}
In order to train our image retrieval pipeline, we need to define a set of relevant positive (visually overlapping) and negative (with no co-visibility) samples for every training query. We build our dataset using hard positive and negative mining.

\noindent\textbf{Positive samples:} To generate positive samples, we randomly sample images from \textit{mixed-conditions} and \textit{overcast-reference}, which share sufficient co-visibility with the query---this method was referred to as 'relaxed inliers' in~\cite{Radenovic2018FinetuningCI}. Randomly sampling within the positive set ensures high variance in viewpoints, while the co-visibility constraint preserves reasonable scale changes.
For \textit{overcast-reference} images, we have access to a 3D representation of the scene. Let $p(I_i)$ be the set of visible 3D points in $I_i$, the positive samples for image $I_q$  with relaxed inliers mining are given by:
\begin{equation}
\resizebox{.48 \textwidth}{!}%
{$
\text{$\mathcal{P}_q$} = \left\{(I_i,c_i): \frac{|p(I_i)\cap p(I_q)|}{|p(I_q)|}>
t_i, i\in\llbracket1,N\rrbracket, i \neq q  \right\}$
}
\end{equation}
We use a value of $t_i=0.6$.

\noindent\textbf{Condition-balanced positive samples:} For images in \textit{mixed-conditions}, we do not have access to a 3D representation of the scene. Thus, we define the set of positives by randomly sampling within poses that fall under translation and orientation distance thresholds.
Let $(R_i, T_i)$ be the absolute camera pose of $I_i$, the positive samples for image $I_q$ captured under condition $j$ with relaxed inliers mining are given by:
\begin{equation}
\resizebox{.48 \textwidth}{!}%
{$
    \text{$\mathcal{P}_q^j$} = \left\{(I_i,c_i):
    \left\{
    \begin{array}{l}
        dist(R_i, R_j) < t_R \\
        dist(T_i, T_j) < t_T \\
        c_i=j \\
    \end{array}
    \right. ,
    i \in \llbracket 1 , N \rrbracket , i \neq q  \right\}$
}
\end{equation}
We use a value of $t_R=10\degree$ and $t_T=8m$.

To ensure a balanced distribution across capturing conditions within \textit{mixed-conditions}, we sample the same number of positive samples for every condition $j\in\{c_1,...,c_{N_c}\}$. Doing so helps regulating over-represented conditions within the training set. See Figure~\ref{fig:sample} for a example of positive samples obtained with this technique.

Since there are far more images in \textit{overcast-reference} than in each condition of \textit{mixed-conditions}, we only use a subset of 2,000 samples from \textit{overcast-reference}, which we resample at every epoch.

\begin{figure}
\begin{center}
\includegraphics[width=1.0\columnwidth]{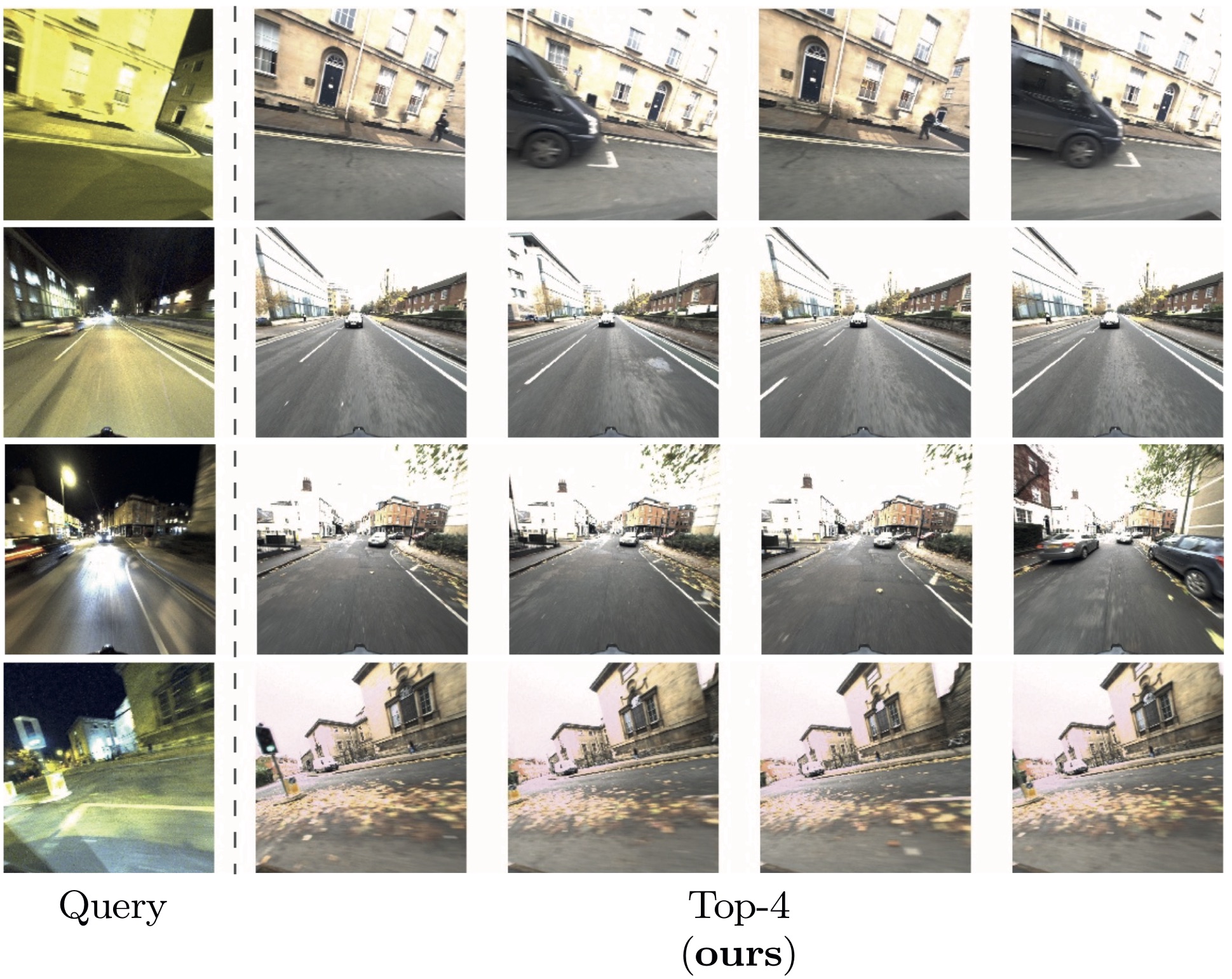}
\end{center}
   \caption{Top-4 retrieval results for nighttime queries. Our method successfully finds query images despite strong illumination and overall appearance changes.}
\label{fig:short}\label{fig:results}
\end{figure}

\noindent\textbf{Negative samples and hard-mining:} We define the set of negative samples as the images which do not present any visual overlap. We pre-compute the negative samples for every query and perform hard-negative mining at every epoch to identify the most challenging images.
\par Our final sample can be written as the t-uple $(I_q, I_{p_1},..,I_{p_P},I_{n_1},...I_{n_N})$, where $P=8, N=8$ is the number of used positive and negative samples per query respectively. We found that applying both mining techniques was crucial to speed up the convergence and make retrieval robust to viewpoint changes. However, by considering images with strong viewpoint changes, we also induce a greater pose approximation when estimating the query pose from top-1 images.

\noindent\textbf{Condition binning:} RobotCar Seasons offers a total of 10 capturing conditions, with the \textit{night} and \textit{night-rain} being the most challenging ones~\cite{sattler:hal-01859660}. While defining a condition-specific backbone for every condition could potentially lead to a more accurate representation pre-processing, this also results in having fewer training samples per sub-branch. Therefore, to increase the number of training samples per backbone, we choose to merge some of the capturing conditions based on their visual similarity. We set $N_C=3$ and combine branches following the repartition presented in Table~\ref{table:condition_binning}.
RobotCar Seasons provides us with very accurate poses, but little training samples per conditions. As a result even when merging conditions, generalizing to unseen places captured under challenging conditions remains particularly hard.

\begin{figure}
\subfigure{\includegraphics[width=0.95\columnwidth]{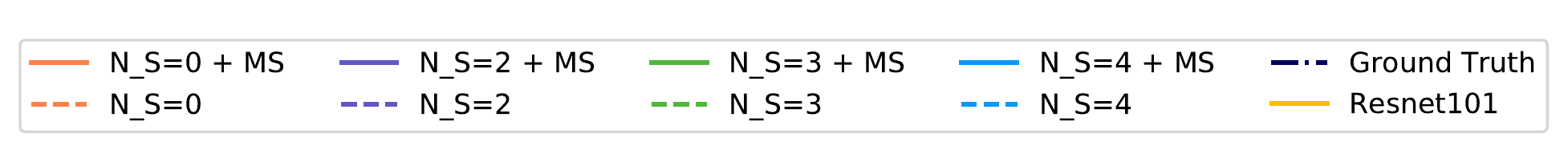}}
\subfigure{\includegraphics[height=1.6in]{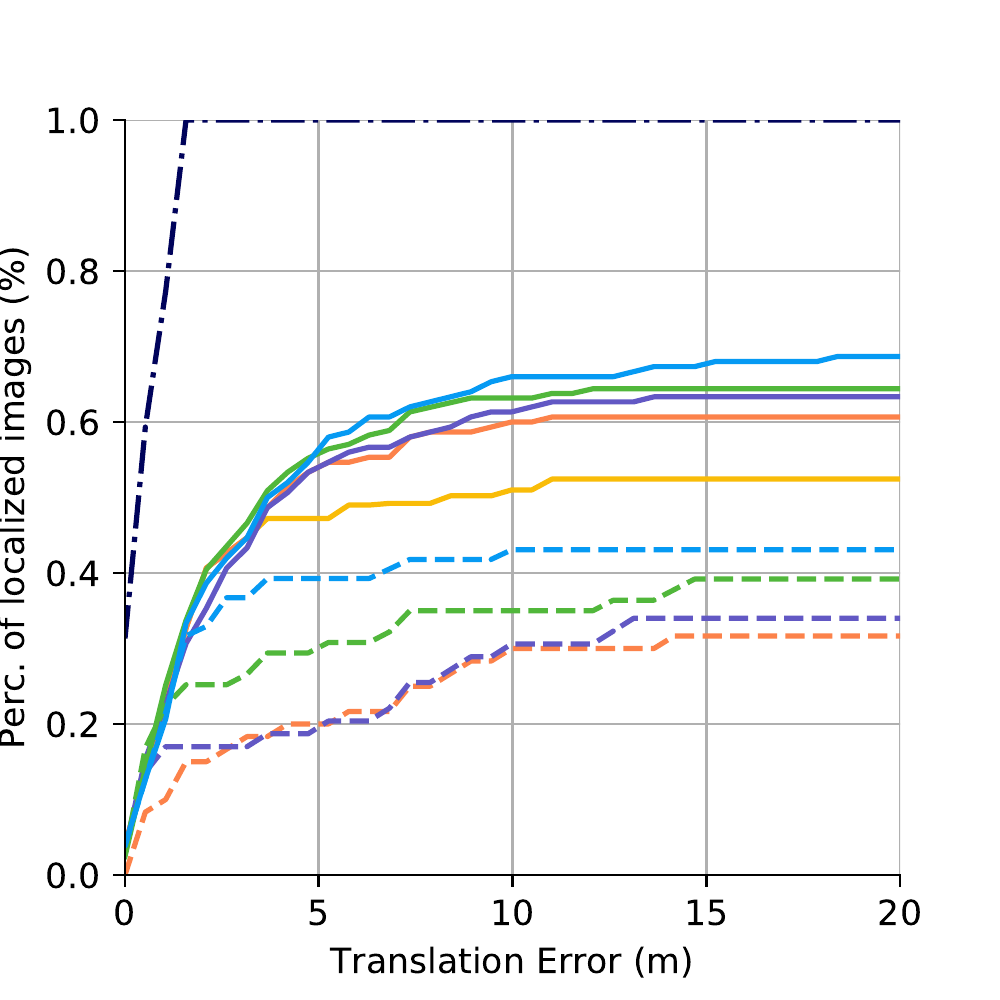}}
\subfigure{\includegraphics[height=1.6in]{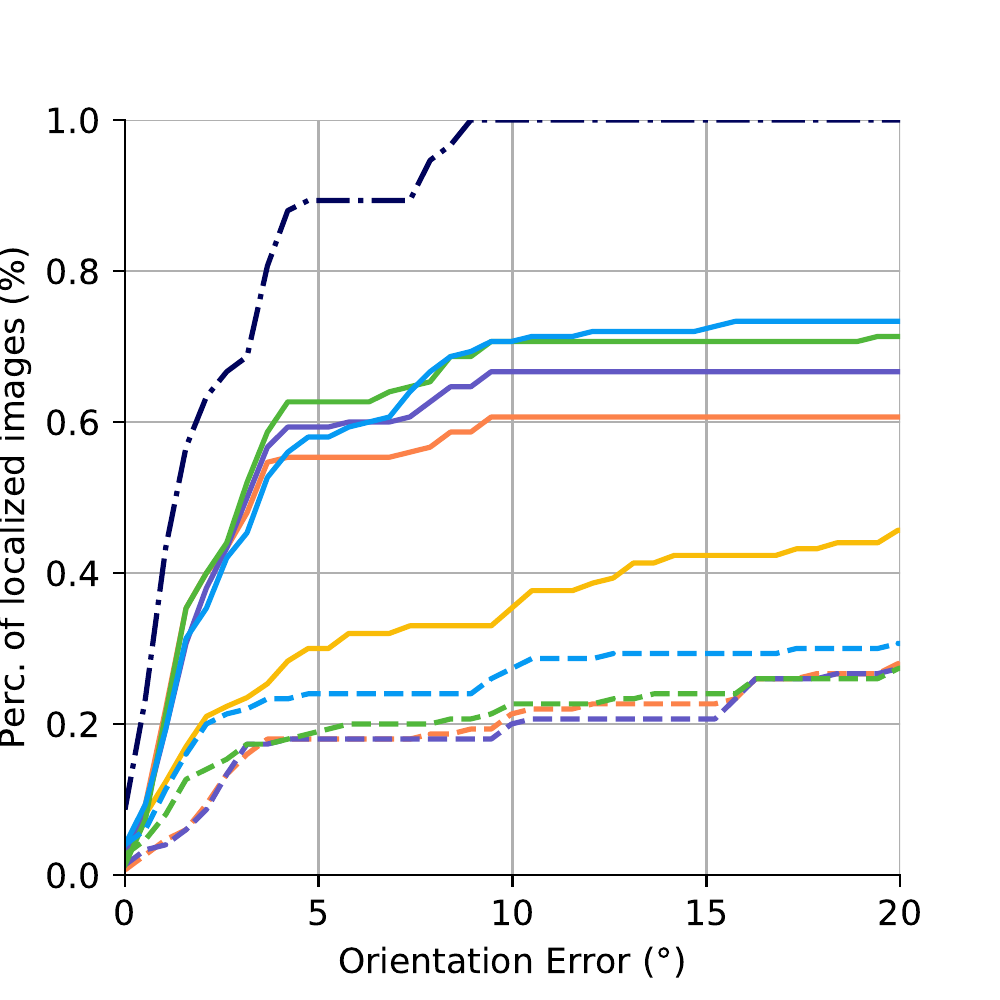}}
\vspace{0.35cm}
\caption{Ablation study of our approach on nighttime images, evaluated on our validation set. We compare the effects of making more resnet blocks condition-specific, as well as the impact of adding pose multi-scale (MS) inference and using a larger backbone (Resnet101). We also include an upper-bound of achievable localization performance computed using the actual nearest neighbour (Ground Truth).}\label{fig:ablation}
\end{figure}

\subsection{Image-retrieval Pipeline}
We train and evaluate our approach on several network architecture variants, followed by some descriptor post-processing steps.

\noindent\textbf{Architectures:} Considering a ResNet-50~\cite{He2016DeepRL} with weights initialized from ImageNet~\cite{imagenet_cvpr09}, coupled with a GeM activation (set with $p_k=3$) followed by $L2$-normalization, let $N_S$ be the number of condition-specific blocks. Each condition-specific branch $f_{c_i}$ for $c_i\in\{c_1,...,c_{N_c}\}$ is therefore made of $N_S$ blocks. We first define a baseline such that $N_S=0$. This results in a traditional siamese network identical to~\cite{Radenovic2018FinetuningCI}, which although popular in image retrieval has not been evaluated so far for the task of visual localization. Then, we train three additional pseudo-siamese networks with $N_S=2, 3, 4$. When $N_S=4$, we end up with $N_c$ separated ResNet-50 being trained jointly. Training is run for 20 epochs, on images rescaled to a resolution of $512 \times512$. Details about the number of parameters used for each variant can be found in Table~\ref{table:params}.\\
\par Lastly, one could argue that when $N_S\neq0$, implicit information on the capturing condition is being propagated until the merging point of the subnetworks. For a fair comparison and to mimic this propagation, we subsequently trained a ResNet-50 with concatenated channels at the entry of each block, containing a discrete label representing the capturing conditions. However, despite our efforts and with weights initializations both from ImageNet~\cite{imagenet_cvpr09} and from Xavier~\cite{Glorot10understandingthe}, we were not able to make this model converge properly and retrieve images correctly.\footnote{Our code will be made publicly available at \url{https://github.com/germain-hug/Improving-Nighttime-Localization}}

\noindent\textbf{Post-processing:} To improve retrieval at inference time, we follow commonly used post-processing steps in image retrieval. First, we learn and apply whitening~\cite{Jgou2012NegativeEA} in a discriminative fashion, as in \cite{Radenovic2018FinetuningCI}. Doing so helps dealing removing noise in descriptors coming from over-represented occurences. In addition, we follow the multi-scale descriptor aggregation proposed by~\cite{Radenovic2018FinetuningCI}, to combine descriptors coming from rescaled inputs using the GeM layer. We feed inputs of size $512 \times 512, 724 \times 724$ and $1024 \times 1024$ pixels.

\begin{figure}
\subfigure{\includegraphics[height=0.32in]{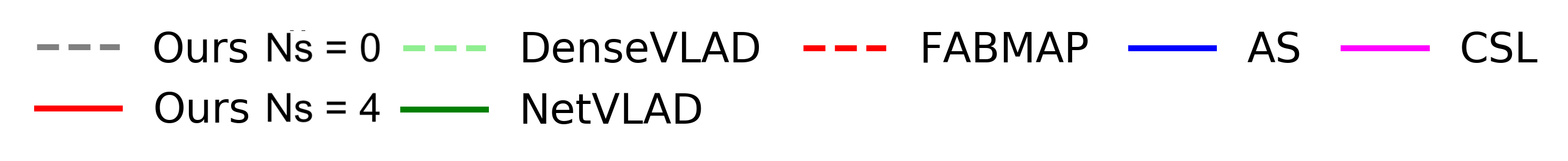}}
~ \vskip\baselineskip
\subfigure{\includegraphics[height=1.2in]{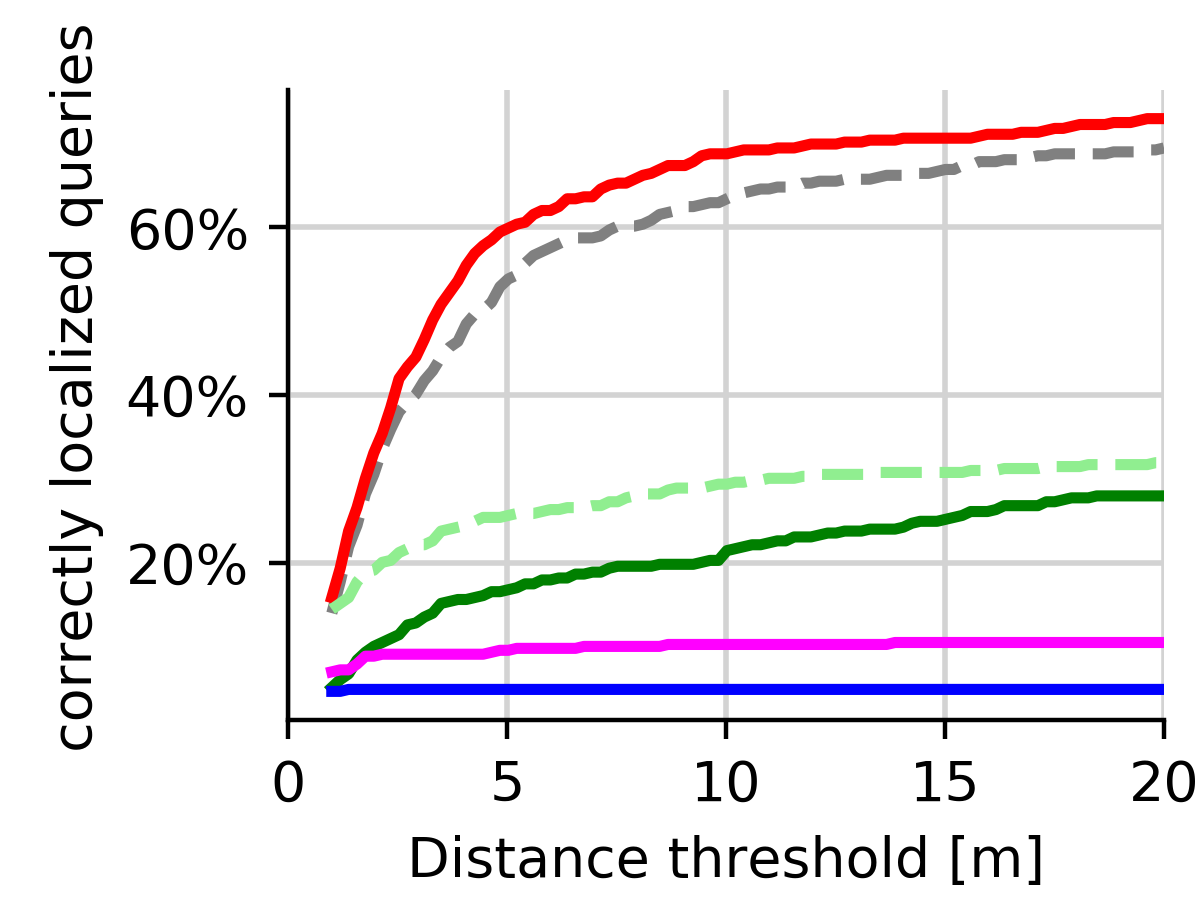}}
\subfigure{\includegraphics[height=1.2in]{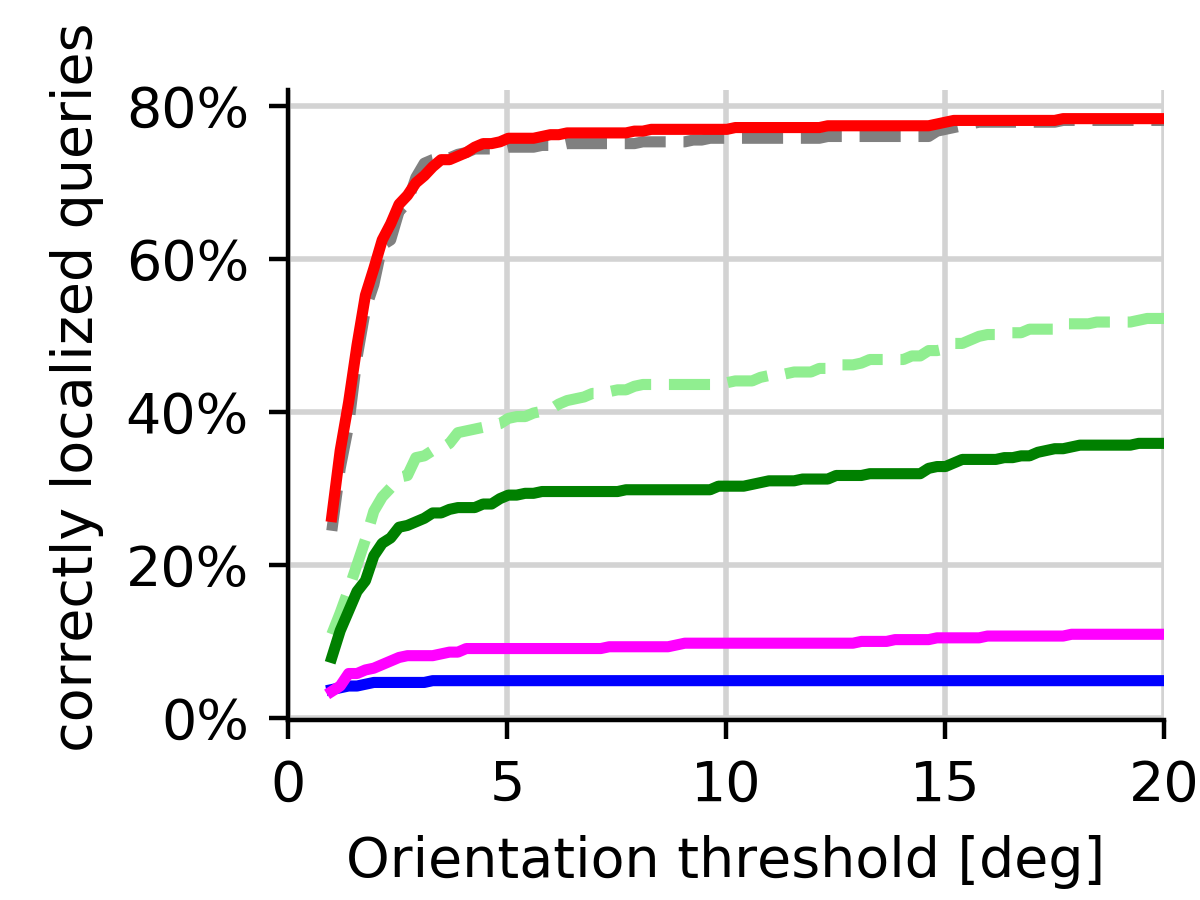}}
\vspace{0.4cm}
\caption{Cumulative distributions over the translation and rotation errors on all nighttime images (\textit{night-all}). Our method  shows significant improvements over existing methods at nighttime for distance error thresholds superior to 3.5$m$.}\label{fig:cumulative}
\label{fig:short}
\end{figure}

\subsection{Evaluation}
We evaluate our approach using the same localization metric as~\cite{sattler:hal-01859660}. Three positional and rotational precision thresholds are defined (0.25m and 2\degree; 0.5m and 5\degree; 5m and 10\degree) and we report the percentage of query images correctly localized within those thresholds. Such binning of localization accuracy is a rather harsh metric for our retrieval-based methods, as they are dependent on the dataset sampling and may not systematically find the nearest image but its neighbors slightly further away. Therefore, we also provide cumulative distributions over the translation and orientation errors under several capturing conditions (see Figure~\ref{fig:cumulative}). We compare ourselves against 2D retrieval-based methods like NetVLAD~\cite{Arandjelovix01072016NetVLADCA}, DenseVLAD~\cite{Torii2015247PR} and FAB-MAP~\cite{Cummins2008FABMAPPL}. In addition, we present results from structure-based methods, which require an additional pre-acquired 3D model of the scene. Such methods include Active Search~\cite{Sattler2017EfficientE}, CSL~\cite{svarm_csl} and a recent variant of Active Search combined with Semantic Matching Consistency~\cite{Toft_2018_ECCV}.

\section{Results}\label{results}

In this section, we present the results of our localization experiments run on the architectures mentioned earlier. We start with an ablation study, followed by a comparison with both retrieval-based and feature-based state-of-the-art methods.

\begin{table}
\ra{1.3}
\begin{center}
\resizebox{\columnwidth}{!}{%
\begin{tabular}{@{} ll *{6}{c@{\hskip0.1in}} @{}}%
& \multicolumn{1}{c}{}
& \multicolumn{6}{c}{\textit{Night-All}}\\
\cmidrule(lr){3-8}
& \multicolumn{1}{c}{}
& \multicolumn{3}{c}{\textit{Night}}
& \multicolumn{3}{c}{\textit{Night-Rain}} \\ %
\toprule
Method
& & \multicolumn{3}{c}{Threshold Accuracy (\%)}
& \multicolumn{3}{c}{Threshold Accuracy (\%)} \\
\cmidrule(lrr){3-5} \cmidrule(lrr){6-8}
& & \makecell{0.25m \\ 2\degree} & \makecell{0.5m \\ 5\degree} & \makecell{5m \\ 10\degree}
&\makecell{0.25m \\ 2\degree} & \makecell{0.5m \\ 5\degree} & \makecell{5m \\ 10\degree}
\\
\midrule
\parbox[t]{6mm}{\multirow{3}{*}{\rotatebox[origin=c]{90}{\makecell{Feature-\\ based}}}} &
CSL~\cite{svarm_csl} & 0.44 & 1.33 & 6.19 & 0.99 & 5.42 & 12.81 \\
& Active Search~\cite{Sattler2017EfficientE} & 0.88 & 2.21 & 3.54 & 0.99 & 3.94 & 6.40 \\
& AS + Sem. Match~\cite{Toft_2018_ECCV} & 10.18(\textasteriskcentered) & 26.54(\textasteriskcentered) & 50.80(\textasteriskcentered) & 7.39(\textasteriskcentered) & 33.50(\textasteriskcentered) & 48.77(\textasteriskcentered) \\
\hline
\parbox[t]{6mm}{\multirow{4}{*}{\rotatebox[origin=c]{90}{\makecell{Retrieval-\\based}}}} &
FAB-MAP~\cite{Cummins2008FABMAPPL} & 0.0 & 0.0 & 0.0 & 0.0 & 0.0 & 0.0\\
& NetVLAD~\cite{Arandjelovix01072016NetVLADCA} & 0.00 & 0.88 & 18.14 & 0.49 & 1.97 & 13.30 \\
& DenseVLAD~\cite{Torii2015247PR} & 0.88 & 4.42 & 24.34 & 2.46 & 5.91 & 25.12\\
& ToDayGAN~\cite{DBLP:journals/corr/abs-1809-09767} & 0.88(\textasteriskcentered) & 9.29(\textasteriskcentered) & 59.29(\textasteriskcentered) & 1.86(\textasteriskcentered) & 12.82(\textasteriskcentered) & 56.41(\textasteriskcentered)\\
\cmidrule(lr){2-8}
& \textbf{Ours ($N_S=0$)} & 0.88 & 4.42 &	50.88 & 2.46 & 6.90 &	54.19 \\
& \textbf{Ours ($N_S=4$)} & 0.88 & 4.42 & 56.19 & 1.97 & 7.88 & 61.58 \\
\bottomrule
\end{tabular}%
}
\end{center}
\caption{Nighttime localization results (values from~\cite{sattler:hal-01859660} accounting for the \textit{mixed-conditions} additional training images) for our approach and state-of-the-art feature-based and retrieval-based methods. (\textasteriskcentered) \textit{\textbf{NB:} AS + Sem. Match~\cite{Toft_2018_ECCV} leverages manually annotated nighttime to train its semantic segmentation network. ToDayGAN~\cite{DBLP:journals/corr/abs-1809-09767} uses 19,998 additional nighttime images to train its GAN network. In comparison, our method requires very few nighttime images to produce these results, and adds no overhead to the overall computational cost of retrieval-based methods.}}\label{table:night}
\end{table}

\subsection{Ablation Study}\label{ablation}
In Figure~\ref{fig:ablation}, we compare the impact of several factors in our architecture. We run this ablation study on a validation set built using 
150 query images taken at nighttime from \textit{mixed-conditions}. First, we evaluate the effect of increasing $N_S$, \ie making more blocks condition-specific. We find that on nighttime images, the network benefits from more computational power dedicated to that condition. In fact, the optimal architecture for nighttime is obtained when all blocks are condition-specific, meaning each condition leverages a full ResNet-50.
\par We also consistently observe that using multi-scale inputs shows strong performance increases on night images. Lastly, we find that using a single larger network such as ResNet-101 damages the results, most likely due to overfitting with the little training data we are provided with. Thus, using a larger architecture naively does not correlate with performance improvement, and our method deals with hard conditions much more effectively.

\subsection{Comparison with state-of-the-art}\label{comparison}

\noindent\textbf{Retrieval-based methods:} Table~\ref{table:night} shows nighttime localization results on the RobotCar Seasons~\cite{sattler:hal-01859660} dataset across all capturing conditions. Our baseline with $N_S=0$ corresponds to the architecture with no condition-specific blocks.
It outperforms both retrieval-based and feature-based approaches at nighttime, for coarse precision thresholds. This first improvement can be attributed to the discriminative power of the GeM~\cite{Radenovic2018FinetuningCI} layer combined with an efficient and condition-balanced positive and negative mining (see Section~\ref{mining}). 

\begin{figure}
\begin{center}
\includegraphics[width=1.0\columnwidth]{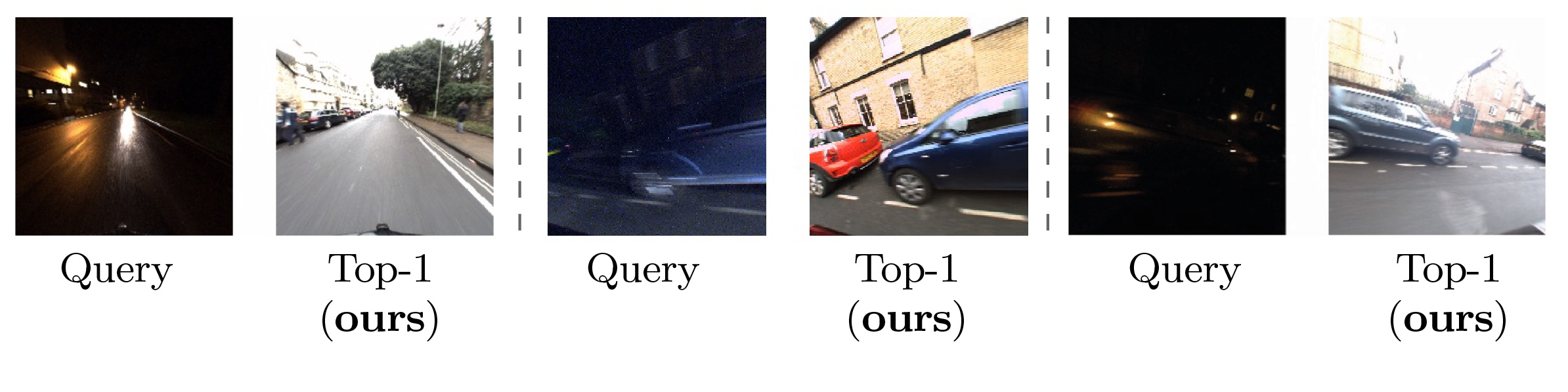}
\end{center}
\vspace{0.1cm}
\caption{Example of failure cases on RobotCar Seasons. Retrieval errors tend to occur when nighttime images are extremely dark and / or present strong blurriness, as well as when the scene is highly occluded.}\label{fig:failures}
\label{fig:short}
\end{figure}

Compared to the baseline, when $N_S=4$ our method further increases nighttime localization for coarse (5m, 10\degree) and medium (0.5m, 5\degree) thresholds by 6.29\% and 0.47\% respectively. Compared to DenseVLAD~\cite{Torii2015247PR}, this increases the results by a factor of 237\% for coarse precision, and by 1200\% compared to AS~\cite{Sattler2017EfficientE}. See Figure~\ref{fig:results} for qualitative results.

The ToDayGAN~\cite{DBLP:journals/corr/abs-1809-09767} method manages to obtain competitive results at nighttime, however it also comes with additional criteria. Indeed, hallucinating a daytime image from a single under-exposed nighttime query is a complex task, which requires a significant amount of additional nighttime images. About 20,000 images were used to train the GAN used in~\cite{DBLP:journals/corr/abs-1809-09767}, unlike our method which uses much fewer images. Besides, as for~\cite{Toft_2018_ECCV} this approach generates a full image at test-time which is a costly operation.\\


\noindent\textbf{Feature-based methods:} In day-time conditions, feature-based methods are especially relevant under high-precision regimes~\cite{sattler:hal-01859660}, where retrieval-based methods are limited by the dataset spatial sampling resolution. However, as observed under \textit{night} and \textit{night-rain} conditions, feature-based methods are prone to failure. Indeed, local descriptors are very sensitive to strong changes in appearance, where global image descriptors such as the ones produced by our approach can efficiently withstand such pertubations.\\

Semantic reasoning as done in~\cite{Toft_2018_ECCV} helps rejecting outliers in feature-based methods, but this approach comes with several requirements. First, this methods needs accurate image segmentation manual labeling on images captured in the same environment. Moreover at test-time, inference of the segmentation map is quite costly compared to retrieval-based methods. Lastly, structure-based methods leverage a 3D model of the scene that needs to be captured and maintained, as discussed in the introduction.

\section{Conclusion and Future Work}
We have presented a novel image-retrieval architecture for visual localization which efficiently and explicitly handles strong appearance variations happening between daytime and nighttime conditions. We showed that our method widely outperforms both state-of-the-art feature-based and retrieval-based methods for coarse nighttime localization. Moreover, our method introduces no additional computational cost at
inference time compared to a standard siamese pipeline.

\begin{table}
\ra{1.3}
\begin{center}
\resizebox{\columnwidth}{!}{
\begin{tabular}{@{}llccc@{}}
\toprule
$N_S$ & $N_A$ & \makecell{Condition-\\Agnostic\\Parameters}  & \makecell{Condition-\\Specific\\Parameters} & \makecell{Total Network\\Parameters} \\
\midrule
$N_S=0$	& $N_A=4$ & 23,508,032 & 0 & 23,508,032\\
$N_S=1$	& $N_A=3$ & 23,282,688 & 225,344 & 23,282,688 + $N_c$ $\times$ 225,344\\
$N_S=2$	& $N_A=2$ & 22,063,104 & 1,444,928& 22,063,104 + $N_c$ $\times$ 1,444,928\\
$N_S=3$	& $N_A=1$ & 14,964,736 & 8,543,296& 14,964,736 + $N_c$ $\times$ 8,543,296\\
$N_S=4$	& $N_A=0$ & 0 & 23,508,032& $N_c$ $\times$ 23,508,032\\
\bottomrule
\end{tabular}}
\end{center}
\caption{Per-branch parameters repartition with respect to the number of condition-specific blocks $N_S$, condition-agnostic blocks $N_A$, and number of discrete conditions $N_c$ on a ResNet-50~\cite{He2016DeepRL} network.
At inference time, the computational cost is independent of $N_S$, $N_A$ or $N_c$ and remains identical to that of a ResNet-50.}\label{table:params}
\end{table}

Improving the accuracy of the predicted pose of retrieval-based methods is therefore the next logical step, and will require being able to align the query image with the retrieved reference image under different conditions.

The performance of retrieval-based localization methods is also tightly linked with the amount of training data available. It should be noted that our approach is complementary to recent works using GANs to generate artificial training images.  One could for example train our method with artificial nighttime samples generated from daytimes sampling, which should be much easier to do than generating daytime samples from nighttime images.

\section*{Acknowledgement}
This project has received funding from the Bosch Research Foundation (Bosch Forschungsstiftung). We gratefully acknowledge the support of NVIDIA Corporation with the donation of the Titan Xp GPU used for this research. The authors would also like to thank Torsten Sattler for providing support and evaluation tools with the RobotCar Seasons dataset~\cite{sattler:hal-01859660}. Vincent Lepetit is a senior member of the \emph{Institut Universitaire de France}~(IUF).


{\small
\bibliographystyle{ieee}
\bibliography{refs}
}

\end{document}